# Towards a Workbench for Acquisition of Domain Knowledge from Natural Language

Andrei Mikheev and Steven Finch

HCRC Language Technology Group
University of Edinburgh
2 Buccleuch Place, Edinburgh EH8 9LW, Scotland, UK
E-mail: Andrei.Mikheev@ed.ac.uk

## Abstract

In this paper we describe an architecture and functionality of main components of a workbench for an acquisition of domain knowledge from large text corpora. The workbench supports an incremental process of corpus analysis starting from a rough automatic extraction and organization of lexico-semantic regularities and ending with a computer supported analysis of extracted data and a semi-automatic refinement of obtained hypotheses. For doing this the workbench employs methods from computational linguistics, information retrieval and knowledge engineering. Although the workbench is currently under implementation some of its components are already implemented and their performance is illustrated with samples from engineering for a medical domain.

## 1 Introduction

One of the standard methods for the extraction of domain knowledge (or domain schema in another terminology) from texts is known as Distributional Analysis (Hirshman 1986). It is based on the identification of the sublanguage specific co-occurrence properties of words in the syntactic relations in which they occur in the texts. These co-occurrence properties indicate important semantic characteristics of the domain: classes of objects and their hierarchical inclusion, properties of these classes, relations among them, lexico-semantic patterns for referring to certain conceptual propositions, etc. This knowledge about domain in the form it is extracted is not quite suitable to be included into the knowledge base and require a post-processing of the linguistically trained knowledge engineer. This is known as a conceptual analysis of the acquired lingistic data. In general all this is a time consuming process and often requires the help of a domain expert. However, it seems to be possible to automate some tasks and facilitate human intervention in many parts using a combination of NLP and statistical techniques for data extraction, type oriented patterns for conceptual characterization of this data and an intuitive user interface.

All these resources are to be put together into a Knowledge Acquisition Workbench (KAWB) which is under development at LTG of the University of Edinburgh. The workbench supports an incremental process of corpus analysis starting from a rough automatic extraction and organization of lexico-semantic regularities and ending with a computer supported analysis of extracted data and a refinement of obtained hypotheses.

## 2 KAW Architecture

The workbench we are aiming at integrates computational tools and a user interface to support phases of data extraction, data analysis and hypotheses refinement. The target domain description consists of words grouped into domain-specific semantic categories which can be further refined into a conceptual type lattice (CTL) and lexico-semantic patterns further refined into conceptual structures as shown elsewhere in the paper. KAW architecture is displayed in figure 1.

**A data extraction module** provides the knowledge engineer with manageable units of lexical data (words, phrases etc.) grouped together according to certain semantically important properties. The data extraction phase can be subdivided into a stage of semantic category identification and a stage of lexico-semantic pattern extraction. Both of these stages complement each other: the discovery of semantic categories allows the system to look for patterns and discovered patterns serve as diagnostic units for further extraction of these categories. Thus both these activities can be applied iteratively until a certain level of precision and coverage is achieved.

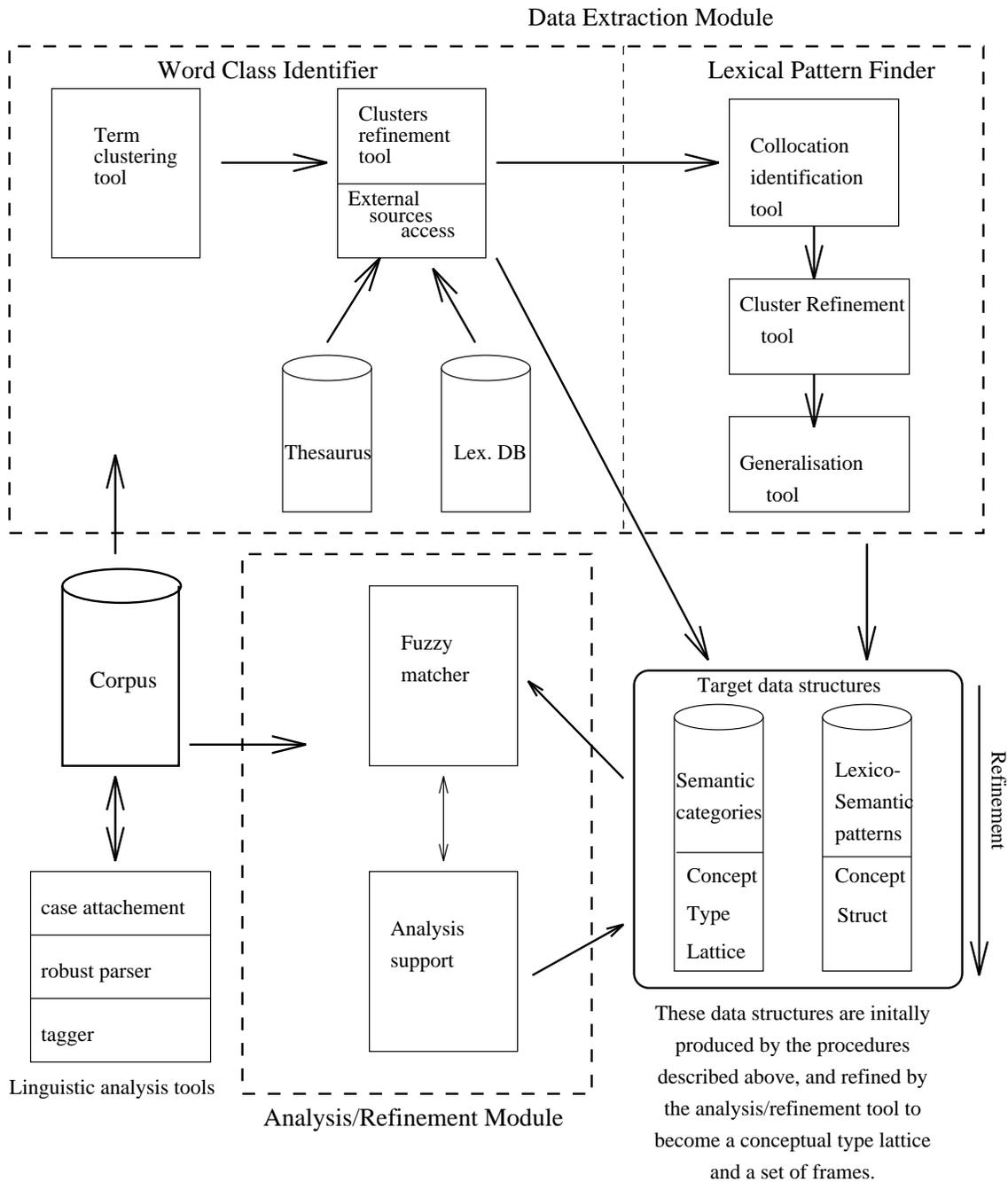

Figure 1: This figure shows main KAWB components and modules and SGML marked data flow between them.

The **word class identification component** encompasses tools for the linguistic annotation of texts, word clustering tools and tools for access to external linguistic and semantic sources like thesauri, machine-readable dictionaries and lexical data bases. *Statistical clustering* can be automatically checked and subcategorized with the help of *external linguistic and semantic sources*.

The **pattern finder component** makes use of phrasal annotations of texts produced by a *general robust partial parser*. First, the corpus is checked for stable phrasal collocations for single words and entire semantic clusters by a special tool - *a collocator*. After collocations are collected another tool - *a generalizer* tries automatically deduce regularities and contract multiple patterns into their general representations. Such patterns are then presented for a conceptual characterization to the knowledge engineer and some predefined generic conceptual structures are suggested for specialization.

The main aim of the **analysis and refinement module** is to uncover and refine structural generalities found in the previous phases. It matches in the text patterns which represent hypotheses of the knowledge engineer, groups together and generalizes the found cases and presents them to the knowledge engineer for a final decision. The matcher evaluates how good a given piece of text matches the pattern and returns matches at various levels of exactness.

If modules are to communicate flexibly then an inter-module information representation format needs to be specified. Standard Generalized Markup Language (Goldfab 1990) is an international standard for marking up text. We use SGML as a way of exchanging information between modules in a knowledge acquisition system, and of storing that information in persistent store when it has been processed.

In the rest of the paper we will embark on a more detailed characterization of tools themselves. We, however, will not present any technical details and suggestions on an actual implementation because the workbench should be able to incorporate different implementations . Some of the tools are already implemented while others still need implementation or reimplementation in terms of the open architecture of the workbench. For an illustration we have used samples from engineering for the cardiac-failure domain using OHSUMED (Hersh 1994) corpus and a corpus of patient discharge summaries ( PDS ) described in Mikheev 1994.

## 3 Linguistic Annotation

The simplest form of linguistic description of the content of a machine-readable document is in the form of a sequence (or a set) of words. More sophisticated linguistic information comes in several forms, all of which may need to be represented if performance in an automatic acquisition of lexical regularities is to be improved.

The NLP module of the KAWB consists of a word tagger (e.g. Kupiec 1993), a specialized partial robust parser and a case attachment module.

The tagger assigns categorial features to words. This is not a straightforward process due to the general lexical ambiguity of any natural language but state-of-the-art taggers do this quite well (more than 97% correctness) using different strategies usually based on an application of Hidden Markov Models (HMMs).

It is well-known that a general text parsing is very fragile and ambiguous by its nature. Syntactic ambiguity can lead to hundreds of parses even for fairly simple sentences. This is clearly inappropriate. However, general and full scale parsing is not required for many tasks of knowledge acquisition but rather a robust identification of certain text segments is needed. Among these segments are compound noun phrases, verb phrases etc. To increase a precision of knowledge extraction in some cases it is quite important to resolve references of pronominal anaphora. At the moment in parsing sentences we are using a temporal expressions recognizer, a noun-phrase recognizer, a simple verb-phrase recognizer and a simple anaphoric binder. This can be further extended to treat other phenomena of natural language, providing that new components are robust and fast.

The parser supplies information to a *case attachment module*. This module using semantically driven role filler expectations for verbs provides a more precise attachment of noun phrases to verbs. To do this we are using ESK - an event and state knowledge base (Whittemore 94) which for more that 700 verbs contains information on thematic roles, semantic types for arguments, expected adjuncts, syntactic information, propositional types, WordNet concept types and sense indices.

## 4 Automatic Precategorization

Semantic clustering of words from an underlying corpora allows the knowledge engineer to find out main semantic categories or types which exist in the domain in question and sort out the lexicon in accordance with these types.

It is important both that information about typology the knowledge engineer adds to the sy-stem is accurate, and that enough information is added. In this regard, the Zipf-Mandelbrot law, which states that the frequency of the $n$th most frequent word in a natural language is (roughly) inversely proportional to $n$. Thus the majority of word tokens appear in a small fraction of the possible word types.

Finch & Chater (1991) show how it is possible to infer a syntactic and semantic classification of a set of words by analyzing how they are used in a very large corpus. This is useful because very large corpora frequently exist for many domains. For example, in the medical domain, the freely available OHSUMED corpus (Hersh 1994) contains some 40 million words of medical texts. We now describe this method for inferring a syntactic and semantic classification of words from scratch.

Firstly, we measure the contexts in which words $w \in W$ occur, and define a statistically motivated similarity measure between contexts of occurrence of words to infer a similarity between words, $d(w_1, w_2), w_1, w_2 \in W$. In our case the context is defined to be a vector of word bigram statistics across the corpus for one and two words to the left and right, thus representing each word to be classified by a vector of bigram statistics. Then we apply a classification procedure to produce a hierarchical single link clustering (or dendrogram) (Sokal & Sneath, 1963) of words which we use as a basis for further classification. If this technique (as more fully described in Finch 1993) is applied to the OHSUMED corpus, some of the structure which is uncovered is displayed in figure 2. This figure displays part of a 3,000 word dendrogram which can then be "cut" at an appropriate level to form a set of disjoint classes which can then be ordered according to their frequency of occurrence.

This gives the knowledge engineer a means to quickly and relatively accurately classify the most frequent vocabulary used in a particular domain.

## 5 Lexico-Semantic Pattren Acquisition

Lexico-semantic patterns are structures where linguistic entries, semantic types and entire lexico-semantic patterns can be used in combinations to denote certain conceptual propositions of the underlying domain and cover certain sequences of words in the text. Linguistic entries can be words, phrases and linguistic types, for example: "of" - word, <NP head = "infarction"> - a noun phrase with the head-word "infarction", <SYNT type = N> - a noun etc. Patterns themselves are the basis for induction of conceptual structures.

An example of a correspondence of many phrases to one lexico-semantic pattern is shown in figure 3. This pattern covers all strings which have a reference to a person followed by one of the listed verbs in any form followed by a compound noun with the head "infarction" and followed by a date expression. In this pattern $PERSON and $DATE are patterns themselves and all other constituents are linguistic entries. If instead of "infarction" we use a type [DISEASE] we can achieve even broader coverage. Also note that the $DATE constituent is optional which is expressed by "?".

A conceptual structure which corresponds to the pattern adds more implicit information to the pattern. For instance, it states explicitly that a body component which is a location of a disease belongs to the person who is an experiencer of that disease:

[@infarction:∀]
    →(is-a)→[@disease]
    →(expr)→[@person:y]
    →(loc)→[@body-COMP]←(has)←[@person:*y]
    →(cul)→[@time-point]

From the NL Processing point of view lexico-semantic patterns provide a way for going about without the definition of a general semantics for every word in the corpus. Many commonsense words take their particular meaning only in a context of domain categories and this can be expressed by means of lexico-semantic patterns.

### 5.1 Collocator

The collocator or multi-word term extraction module finds in the corpus significant co-occurrence of lexical items (words and phrases) which constitute terminology. Identified by the robust partial parser noun and verb groups which include domain semantic categories elicited at the precategorization phase are collected together with frequencies of their appearance in the corpus. Phrases are filtered through a list of general purpose words which is constructed separately for every new domain. Phrases which occur more often than a threshold computed using Zipf-Mandelbrot law are saved for post-analysis. Other phrases are decomposed into constituents for recalculation of saved phrase weights as described in Mikheev 91.

Many terms include other terms as their components. This surface lexical structure corresponds to semantic relations between concepts represented by these terms. To uncover term inclusion the system scans the term bank and replaces each entry of a term which currently in focus with its number. Figure 4 displays an excerpt from collocations extracted from PDS corpus in the original form and after term inclusion checking.

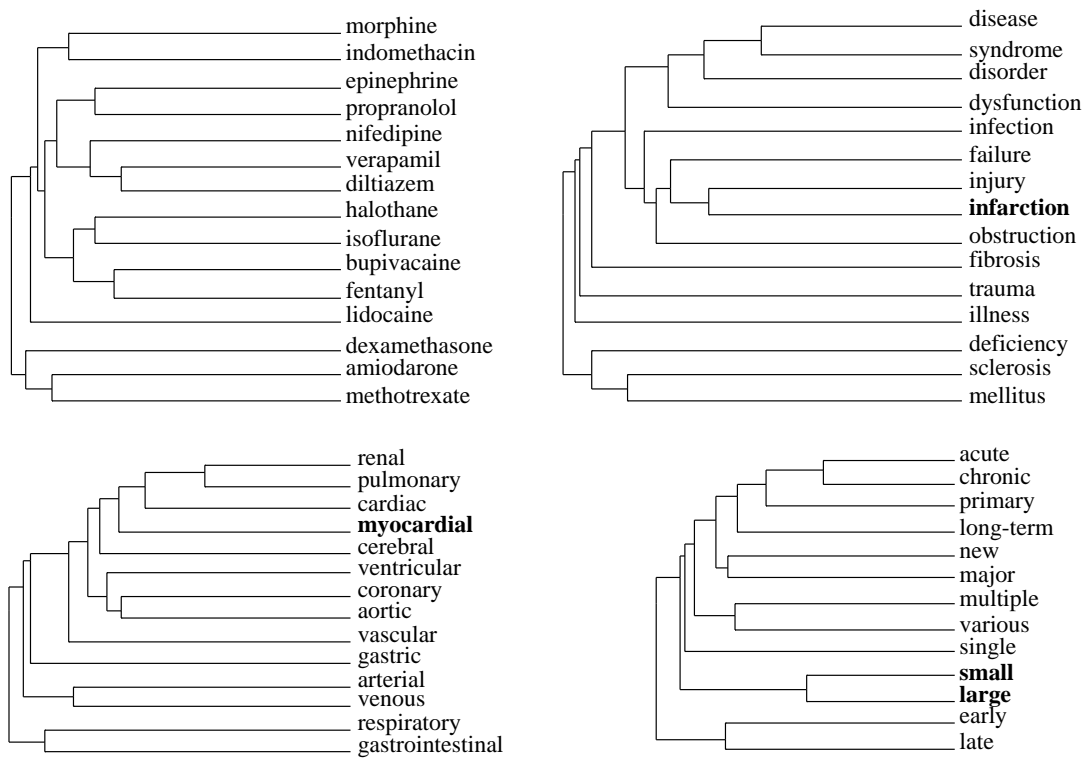

Figure 2: This figure shows four sub-clusters of our hierarchical cluster analysis of the 3,000 most frequent words in the OHSUMED corpus (Hersh 1994). It shows a subcluster of drugs (top left), disease-based nouns (top right), body-part adjectives (lower left), and condition modifying adjectives (lower right).

| He had suffered | an acute myocardial infarction | in 1992. |
| He had | a true posterior myocardial infarction | on 5th of November 1992... |
| She had had | an interior infarction | in 1985. |
| Mr.Mcdool sustained | a small anterior myocardial infarction | in October 92. |
| She developed | an extensive myocardial infarction | |

$PERSON <V head = {suffer, have, sustain, develop} > <NC head = "infarction"> {"on", "in"} $DATE?

Figure 3: This figure shows a correspondence of many phrases to one lexico-semantic pattern.

| Num | Freq | Annotated Phrase |
|---|---|---|
| $136 | 373 | myocardial//BODY-PART infarction//DISEASE |
| $234 | 475 | anterior myocardial//BODY-PART infarction//DISEASE |
| $467 | 550 | inferior myocardial//BODY-PART infarction//DISEASE |
| $1109 | 17 | established inferior myocardial//BODY-PART infarction//DISEASE |
| $1154 | 48 | history//INFORMATION of ischaemic heart//BODY-PART disease//DISEASE |
| $2574 | 21 | history//INFORMATION of an anterior myocardial//BODY-PART infarction//DISEASE |
| $2974 | 23 | moderately severe stenosis//DISEASE |
| $2980 | 46 | aortic//BODY-PART valve//BODY-PART stenosis//DISEASE |
| $3004 | 79 | stenosis//DISEASE in the right coronary//BODY-PART artery//BODY-PART |

Figure 4: This figure shows an excerpt from collocations extracted from PDS corpus and the result of term inclusion checking.

## 5.2 Inner Context Categorization

The major part of the terminology is usually represented by nouns or nominalizations. Such terms usually have a particular set of modifiers which represent different properties. The inner context categorization is started with extraction of compound nouns from collected by the collocator noun phrases.

Semantic categories for many adjectival modifiers extracted at the word clustering phase are too general if any, but collected collocations and external lexical sources as, for example, WordNet can be used.

First, we can sort terms with the same head-word by length. For example, for the type INFARCTION the systems sorts terms as follows:

myocardial infarction, old infarction, acute infarction acute myocardial infarction, anterior myocardial infarction...

further anterior myocardial infarction...

Then we separate pure adjectival modifiers from adjectivized nouns:

**infarction** : inferior, old, acute, post, further, anterolateral, lateral, infero-posterior, antero-septal, repeated, significant, large, limited // myocardial, diaphragmatic, subendocardial

**myocardial infarction** : anterior, first, extensive, minor, small, previous, posterior, suspected.

Next we cluster pure adjectival modifiers into groups using synonym-antonym information available in WordNet. However, it is not necessarily the case that related adjectives are stated together in one WordNet entry. Sometimes there is an indirect link between adjectives. Also, since quite often WordNet gives semantically unrelated (in a given domain) adjectives together we use a heuristic rule which says that *if two adjectives are used together in one phrase they don't hold synonymy/antonymy relation.*

The system assumes that if there is at least one word in common in WordNet entries for two different adjectives they can be clustered together. In our example for the type INFARCTION the following clusters were automatically obtained:

**cluster 1**: chronic vs. acute;

**cluster 2**: major, extensive, significant, large, old vs. minor, small, limited;

**cluster 3**: post vs. previous, ensuing;

**cluster 4**: anterior vs. posterior;

**cluster 5**: inferior vs. superior;

**rest**: suspected; lateral; recent; further; repeated;

As we see all clusters look fairly plausible except the single adjective "old" which was misclassified; it stands for a temporal property of an infarction rather than its spreading at a myocardium.

This algorithm is gradually applied to all entries from the term bank and the knowledge engineer is presented with the results. This method was fairly successfully used in our experiment, however, a large-scale evaluation of sense discrimination for constituent words is still needed to be done.

## 5.3 Outer Context Generalizer

The lexico-semantic generalizer is a tool which extracts general lexico-semantic patterns in an empirical, corpus-sensitive manner analogous to that used to automatically extract word class dendrograms. From the multi-word term bank collected by the collocation tool, we derive semantic frames by replacing each content word in each phrase by its semantic category, derived either empirically from the word-level dendrogram in the case of frequent words, or derived from WordNet in the case of less frequent words (as described above). We also part of speech tag every word in the phrase. Therefore, the term "myocardial infarction" might become "BODYPART<adj> DISEASE<noun/s>", as might "gastrointestinal obstruction" or "respiratory failure". Another example might be the assignment of "DISEASE<noun/s> of BODYPART<noun/pl>" to "obstruction of arteries" (function words such as "of" are usually not further subcategorized, since they convey structural information in themselves). Thus we map the term bank to a set of paradigms, and we choose the set of paradigms which appear most frequently for clustering.

Clustering proceeds by mapping words in the corpus to their semantic category (augmented with part-of-speech information), and clustering in the same way as we did for words, except that the context vectors are recorded for the set of frequent semantic paradigms. For infrequent words where the empirical method for finding semantic class can't be applied, the WordNet technique described above is used. When this is done, we get a clustering of short lexico-semantic paradigms .

Once this is achieved, we can again apply the same methodology to find patterns of higher level which include patterns themselves. In our notation, we refer to singe word semantic categories as uppercase labels (which we choose as being descriptive of the class which has been discovered), simple sequences of semantic categories by a preceding "$", and a sequence of sequences by a preceding "$$". These higher level patterns can be clustered in the same way to yield longer semantic sequence paradigms. Figure 5 illustrates generalizations for the types $BODY-PART and $$DISEASE.

| Pattern Structure for $BODY-PART | Examples |
|---|---|
| BODY-PART< *adj* > BODY-PART< *noun/s* > | aortic valve |
| LOCATION< *adj* > BODY-PART< *noun/s* > | left heart |
| LOCATION< *adj* > LOCATION< *adj* > BODY-PART< *noun/s* > | left descending artery |

| Pattern Structure for $$DISEASE | Examples |
|---|---|
| $BODY-PART DISEASE< *noun/s* > | antero-septal myocardial infarction |
| DISEASE< *noun/s* > "in" $DATE | infarction in December 1987 |
| $BODY-PART DISEASE< *noun/s* > "in" $DATE | myocardial infarction in December 1987 |
| DISEASE< *noun/s* > "of" $BODY-PART | occlusion of artery |

Figure 5: This figure shows results of generalization for the types $BODY-PART and $$DISEASE.

### 5.4 Analysis Support Tool

Type oriented analysis is facilitated with generic conceptual structures which are different for different conceptual types (as more fully described in Mikheev & Moens 1994). For example, a type oriented structure for eventualities includes their thematic roles (agent, theme ...), temporal links and properties while a type-oriented structure for objects includes their components, parts, areas and properties. The system recognizes which structure should be used and presents it to the knowledge engineer with optional explanations or a question guided strategy for filling it up.

## 6 Hypotheses Refinement

A fuzzy matcher is a tool which uses a sophisticated pattern-matching language to extract text fragments at various levels of exactness. It matches in the text patterns which represent hypotheses of the knowledge engineer, groups together and generalizes cases which have been discovered and presents them to the knowledge engineer for a final decision.

Patterns themselves can be quite complex constructions which can include strings, words, types, precedence relations and distance specifiers. In the simplest case the knowledge engineer can examine a context for occurrences for a word or a type provided that the type exists in the term bank as represented in figure 6.

More complex patterns can be used for the description of complex groups. For instance, there a request can be made to find all co-occurrences of the type DISEASE with the type BODY-COMPONENT when they are at the same structural group (noun phrase or verb phrase) and the disease is a head of the group:

{[disease].<>[body-component]}

curly brackets impose a context of a structural group, the "." means that the words can be distributed in the group, <> means that the component can be both to the left and to the right, and since the DISEASE is the first element of the pattern it is assumed to be the head. The program matches this pattern into the following entries:

myocardial infarction, infarction of myocardium, stenosis at the origin of left coronary artery...

To be powerful enough for our purposes this pattern language should be quite complex and it is important to provide an easy way for specification of such patterns with a question-guided process.

## 7 External Sources Access

Already existing lexical databases are an important source of information about constituent words of domain texts. KAWB provides generic facilities for access to such linguistic sources. For each source a converter which transforms source information into SGML marked data, which then can be used in the workbench, should be written.

For some domains there already exist terminological banks available on-line. These banks vary in their linguistic coverage - some list all possible forms (singular, plural etc.) for terms while others just a canonical one, and in a conceptual coverage - some provide an extensive set of different relations among terms (concepts) others just a subsumption hierarchical inclusion. In our implementation we used Unified Medical Language System (UMLS) and WordNet (Beckwith *et al* 1990) - a publicly available lexical database, however we haven't provided the generic support for an abstract thesaurus yet.

## 8 Conclusion

The workbench outlined in this paper encompasses a number of tools which facilitate different stages of knowledge extraction, analysis and refinement based on corpus processing paradigm. These tools are integrated into a coherent workbench with a common inter-module data flow interface based on SGML. Thus the workbench can easily integrate new tools and upgrade existing ones.

| | | |
|---|---|---|
| developed an anterior myocardial | infarction | from which |
| an established inferior myocardial | infarction | . The |
| an acute inferior myocardial | infarction | with CHB |
| subsequent episodes of unstable | angina | including an |
| he has experienced unstable | angina | and was |

Figure 6: This figure shows an excerpt from a search for the type DISEASE with a distance four to the left and two to the right.

The general approach to knowledge acquisition supported by the workbench is a combination of methods used in knowledge engineering, information retrieval and computational linguistics.

## References

Beckwith, R., C. Fellbaum, D. Gross and G. A. Miller (1990) WordNet: A lexical database organized on psycholinguistic principles. CSL Report 42, Cognitive Science Laboratory, Princeton University, Princeton.

Cutting, D., J. Kupiec, J. Pedersen and P. Sibun (1993) Beta test version of the Xerox tagger. Xerox Palo Alto Reseach Center, Palo Alto, Ca.

Finch, S. and N. Chater (1991) A hybrid approch to learning syntactic categories. *AISB Quarterly* **8**(4), 35–41.

Finch, S. P. (1993) *Finding Structure in Language*. PhD thesis, Centre for Cognitive Science, University of Edinburgh, Edinburgh.

Goldfarb, C. F. (1990) *The SGML Handbook*. Oxford: Clarendon Press.

Health, U S. D.of (1993) *UMLS Knowledge Sources*. Washington: National Library of Medicine.

Hersh, W. (1994) An interactive retrieval evaluation and a new large test collection for research. In W. B. Croft and C. J. van Rijsbergen, eds., *Proceedings of the 17th Annual International Conference onResearch and Development in Information Retrieval*, pp. 192–202.

Hirschman, L. (1986) Discovering sublanguage structures. In R. Grishman and R. Kittredge, eds., *Analyzing Language in Restricted Domains: Sublanguage Description and Processing*, pp. 211–234. Hillsdale, N.J.: Lawrence Erlbaum Associates.

Mikheev, A. and M. Moens (1994) Acquiring and Representing Background Knowledge for a NLP System. In *Proceedings of the AAAI Fall Symposium*.

Mikheev, A. (1991) *A cognitive system for conceptual knowledge extraction from NL texts*. PhD thesis, Computer Science, Moscow Institute for Radio-Engineering and Automation, Moscow.

Sokal, R. R. and P. H. A. Sneath (1963) *Principles of Numerical Taxonomy*. San Fransisco: W. H. Freeman.

Whittemore, G. and J. Hicks (1994) ESK: Event and state knowledge base. In *AAAI Fall Symposium*.